\documentclass[onecolumn]{article}
\usepackage{arxiv}  %
\usepackage[T1]{fontenc}

\usepackage[square,numbers]{natbib}

\usepackage{multirow}
\usepackage{graphicx}
\usepackage{times}

\graphicspath{ {./} }
\newcommand{\method}{SUHMo}

\usepackage{booktabs}
\RequirePackage{fontawesome}

\usepackage{enumitem}
\setlist{leftmargin=*,parsep=0.3em,itemsep=0cm,topsep=0.3em,partopsep=0.2em}

\usepackage{amsmath,amsfonts,bm}

\def\eqref#1{equation~\ref{#1}}

\def\1{\bm{1}}

\DeclareMathAlphabet{\mathsfit}{\encodingdefault}{\sfdefault}{m}{sl}
\SetMathAlphabet{\mathsfit}{bold}{\encodingdefault}{\sfdefault}{bx}{n}

\usepackage[pdfencoding=auto, pdfusetitle]{hyperref}
\hypersetup{
  hidelinks,
  urlcolor=red,
  pdftitle={Autoregressive GAN for Semantic Unconditional Head Motion Generation},
  pdfauthor={Louis Airale, Xavier Alameda-Pined, St\'ephane Lathuili\`ere, Dominique Vaufreydaz},
  pdfkeywords={KEYWORD1, KEYWORD2, KEYWORD3},
}

\title{Autoregressive GAN for Semantic Unconditional Head Motion Generation}

\author{Louis Airale$^{1,2}$, \href{https://xavirema.eu/}{Xavier Alameda-Pineda\textsuperscript{2, \tiny \faExternalLink}}, \href{https://stelat.eu/}{St\'ephane Lathuili\`ere\textsuperscript{3, \tiny \faExternalLink}}, \href{https://research.vaufreydaz.org/}{Dominique Vaufreydaz\textsuperscript{1, \tiny \faExternalLink}}
\\
{$^1$ Univ. Grenoble Alpes, CNRS, Grenoble INP, LIG, 38000 Grenoble, France}\\ %
{$^2$ Univ. Grenoble Alpes, Inria, CNRS, Grenoble INP, LJK, 38000 Grenoble, France}\\
{$^3$ LTCI, T\'{e}l\'{e}com Paris, Institut Polytechnique de Paris, France}\\
}

\begin{document}

\setcounter{footnote}{0}

\maketitle

\begin{abstract}

In this work, we address the task of unconditional head motion generation to animate still human faces in a low-dimensional semantic space from a single reference pose. 
Different from traditional audio-conditioned talking head generation that seldom puts emphasis on realistic head motions, we devise a GAN-based architecture that learns to synthesize rich head motion sequences over long duration while maintaining low error accumulation levels.
In particular, the autoregressive generation of incremental outputs ensures smooth trajectories, while a multi-scale discriminator on input pairs drives generation toward better handling of high- and low-frequency signals and less mode collapse.
We experimentally demonstrate the relevance of the proposed method and show its superiority compared to models that attained state-of-the-art performances on similar tasks. 

\vspace{1em}

\end{abstract}

\section{Introduction}

Talking head generation refers to the task of animating a human face generally using a single reference image, an audio clip, and possible additional conditioning signals such as emotional state or exemplar pose dynamics~\cite{taylor2017deep,suwajanakorn2017synthesizing, karras2017audio,yi2020audio,zhou2021pose}.
Different from face reenactment where a driving video clip is provided, in talking head generation the head pose, facial animation, and lip synchronization need to be inferred from other modalities. %
To tackle the difficulty of handling both facial dynamics and photorealism directly in the image space, a predominant line of research generates dynamics in a lower dimensional space~\cite{villegas2017learning}.
Those representations comprise supervised facial landmarks~\cite{chen2019hierarchical,zhou2020makelttalk}, 3D mesh~\cite{fan2022faceformer} or unsupervised keypoints~\cite{Wang2021Audio2HeadAO,wang2022one}, and following the designation of \textit{high level semantics} used in~\cite{villegas2017learning}, we  refer to this space as the \textit{semantic space}.

Although several works achieved compelling results in lip-syncing and realistic rendering, generating natural head motions has, until recently, consistently received less attention.
In the lack of a driving audio signal, it is yet crucial for the synthesis model to produce natural and diverse head motions.
This is relevant in applications where no audio signal is available, e.g.\ when animating background characters in a scene or a video game.
In this unconditional generation setting, the focus shifts from audio-visual synchrony toward long-term consistency of unguided generation, which is known to be particularly challenging~\cite{vondrick2016generating}.
Tackling this problem can also be beneficial for audio-conditioned talking head synthesis, as it fosters fine handling of head dynamics.
In this work, we address the task of unconditional head motion sequence generation, i.e. synthesizing head pose and facial expression given a single reference pose and no audio driving signal.
We do so using a low dimensional \textit{semantic} representation, namely 2D facial landmarks, that facilitates the manipulation of head dynamics and can easily be mapped back to the image domain~\cite{zakharov2019few,zakharov2020fast,zhao2021sparse,meshry2021learned}.

Continuous sequence prediction problems such as head motion generation have previously been addressed by producing first order derivatives, e.g. instantaneous velocities instead of positions, for different tasks such as human trajectory prediction~\cite{gupta2018social} or human pose generation~\cite{lin2018human,kundu2019bihmp}.
One advantage of generating residual quantities is that it can allow using shallower networks~\cite{martinez2017human}.
Second, it is conveniently modeled by autoregressive generators which provide an inductive bias for cumulative sum operations~\cite{morrison2022chunked}.
However, autoregressive models may accumulate error, or alternatively end up generating average values over time when trained with a mean squared error loss~\cite{martinez2017human}, which advocates for the use of other loss functions.
We hereby introduce an adversarial framework to tackle head motion generation as an autoregressive velocity prediction problem, which to the best of our knowledge has never been done before in talking head generation.
To that end, we carefully designed our discriminator network by taking advantage of the specificity of head motion data.
Head motion dynamics are structured data composed of temporal patterns that evolve over varied timescales.
Previous works have addressed structured data generation with discriminator networks operating on receptive fields of different sizes~\cite{wang2018high,kumar2019melgan} or on local windows, enabling a better representation of high-frequency components~\cite{isola2017image}.
We build on this knowledge and use a multi-scale, window-based discriminator, but noticeably implement it in a single network, allowing us to flexibly incorporate any new resolution.
Last, to mitigate mode collapse we follow~\citet{lin2018pacgan} and provide input pairs to the discriminator network, but also produce samples together in the generator.
While this does not change the optimization objective, it brings a significant performance boost for a limited additional overhead.
The proposed GAN architecture, labeled Semantic Unconditional Head Motion or \method{}, allows for long-term head motion synthesis, and experiments confirm its proficiency against a diversity of models and baselines.\footnote{Source code and animated examples can be found at: \href{https://github.com/LouisBearing/UnconditionalHeadMotion}{https://github.com/LouisBearing/UnconditionalHeadMotion}.}

The contributions of this research work are:
\begin{itemize}
    \item An autoregressive GAN framework for unconditional head motion generation in the 2d-landmarks domain, able to mitigate error accumulation over long sequences, even extending the duration of training sequences,
    \item A training methodology that can be generalized over diverse architectures, for which we detail two implementations based on LSTM and Transformers,
    \item Extensive experiments showing that the proposed \method{} method surpasses competitive methods from closely related tasks on two benchmark datasets.
\end{itemize}

\section{Related work}

\subsection{Talking head generation} 
Talking head generation aims at syncing driving audio with head motions, and has seen tremendous recent progress~\cite{chen2019hierarchical,zhou2019talking,vougioukas2020realistic,zhou2020makelttalk,wang2022one,stypulkowski2023diffused}.
Although early identified as a key component for faithful face animation~\cite{greenwood2018joint}, the prediction of head pose and facial expression beyond lip region has been noticeably less investigated, in favor of the use of a driving head motion sequence~\cite{yu2020multimodal,zhou2021pose,ji2022eamm}.
As it is a one-to-many mapping, learning to generate head motion from audio is challenging, and the usual mean squared error loss typically produces static average poses.
Successful attempts at handling head poses include~\cite{chen2020talking,yi2020audio,zhou2020makelttalk}, although the range of achieved motions remains limited.
Recently,~\citet{Wang2021Audio2HeadAO} presented natural-looking results with head pose and face expression produced in a sparse keypoints manifold by two separately trained modules, and further extended their work in~\citet{wang2022one}.
In comparison, our model generates all semantic data in a single module, learning possible correlations between pose and expression, and uses an autoregressive formulation to enforce temporal consistency.

\subsection{Deep continuous autoregressive models} 
Autoregressive models are ubiquitous in sequence modeling, as they enable strong temporal consistency thanks to the explicit relation between consecutive outputs.
In the context of deep continuous sequence prediction, autoregressive models proved powerful in as diverse domains as waveform synthesis~\cite{kumar2019melgan}, human trajectory prediction in a crowd~\cite{gupta2018social}, or human motion prediction~\cite{martinez2017human,lin2018human,kundu2019bihmp,aliakbarian2021contextually}.
Surprisingly, the talking face generation literature is much sparser on this subject,~\citet{fan2022faceformer} presenting one of the few autoregressive talking head generation architectures, but they do not attempt to generate head motions.
Different from previous works, we leverage the potential of autoregressive models to produce smooth and realistic head motions.

\subsection{Multi-scale generative adversarial networks} 
Uncovering multiple patterns with GANs was first addressed in~\citet{isola2017image} where the authors introduced a discriminator network taking image patches as input to enhance high spatial frequency components.
In~\citet{wang2018high}, an output image pyramid is processed by several discriminators that operate on decreased resolutions and larger receptive fields, driving the generator network to produce realistic patterns at different scales.
The multi-scale discriminator has then been extended to sequence generation tasks~\cite{lin2018human,kumar2019melgan}.
An interesting aspect of the latter discriminator architectures is that they combine multi-scale with window-based evaluations in a 1D equivalent of PatchGAN~\cite{isola2017image}, and benefit from the advantages of processing short windows, such as a lighter architecture and faster inference.
Our window-based multi-scale discriminator follows that of~\citet{airale2022socialinteractiongan}, which has the additional advantage of being very flexible regarding the evaluation scales, for a fixed number of parameters.

\subsection{Mode collapse mitigation} 
Mode collapse reduction methods in GANs have comprised efforts towards better optimization procedures~\cite{pmlr-v70-arjovsky17a}, generation space regularization~\cite{che2016mode}, or forcing the network to account for the noise vector~\cite{chen2016infogan}, among a rich body of literature.
\citet{lin2018pacgan} proposed an intuitive way of driving the generator to produce diverse outputs by feeding the discriminator with several input samples.
We extend this framework by generating two inputs \textit{together}, which provided better results while leaving the optimization objective unchanged.

\section{Autoregressive unconditional head motion generation}
\label{sec_p3}

In this section, we formally define the unconditional head motion generation task and the key components of our learning framework. Given a set of facial landmarks $x_0$ representing a face in an initial pose, we seek to generate a sequence $x_{1:T}=(x_1, \ldots, x_T)$ of arbitrary length $T$ such that the probability distributions of the generated  and the ground truth data, $p_{\textsc{g}}$ and $p_{\text{data}}$, match:
\begin{equation}
    p_{\textsc{g}}(x_{0:T}) = p_{\text{data}}(x_{0:T}), \quad\forall x_{0:T}
\end{equation}

\begin{figure*}
\includegraphics[width=1.0\linewidth]{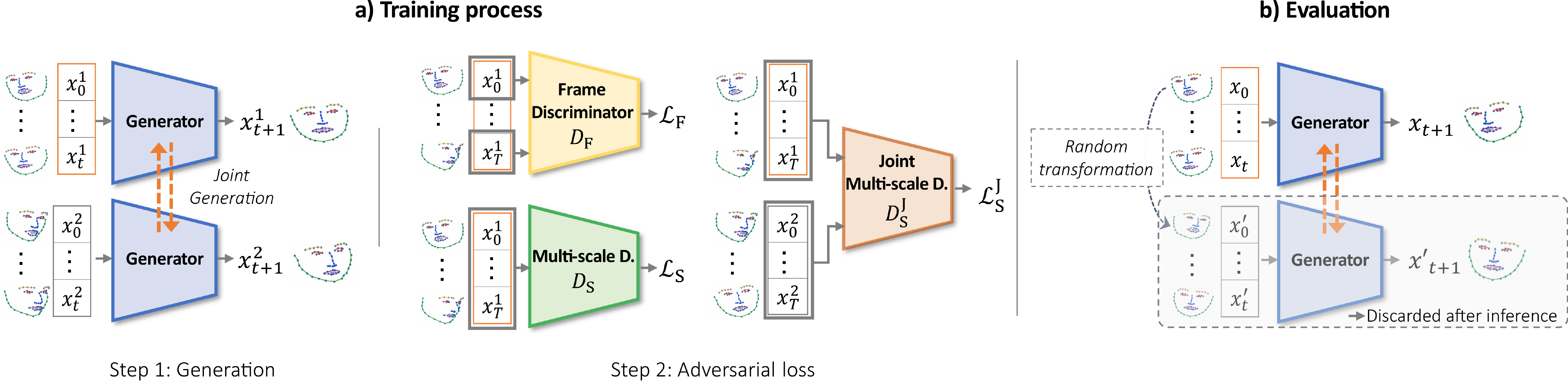}
\centering
\caption{Overview of \method{} training process. The autoregressive generator produces pairs of outputs, that are evaluated by three discriminator networks. At test time, the second sample is replaced by a transformed version of the reference pose.}
\label{fig:overview}
\end{figure*}

We hereafter describe our adversarial architecture to address this problem, an overview of which can be found in Figure~\ref{fig:overview}.
Its main components include the autoregressive generator, described in Section~\ref{subsec:generator}, and the multi-scale sequence discriminator, presented in Section~\ref{subsec:wbmsd}. 
As an attempt to mitigate the potential negative impact of mode collapse, we design our architecture to learn to generate and discriminate joint probability distributions, as explained in Section~\ref{subsec:joint}.
The overall loss function is presented in Section~\ref{subsec:arch_loss}.
Finally, in Section~\ref{subsec:implem} we propose two implementations of our method to stress its generalizibility.

\subsection{Autoregressive velocity generation}
\label{subsec:generator}

We implement our generator network $G$ as an autoregressive function of past landmark positions, that at each time steps provides the instantaneous velocity:
\begin{equation}
    x_t = x_{t-1} + G(x_{0:t-1})
\end{equation}
Working with velocities ensures smooth transitions between subsequent time steps but also enables simpler model architectures~\cite{martinez2017human} and provides a convenient way to take advantage of the inherent potential of autoregressive models to represent cumulative sums~\cite{morrison2022chunked}.
On the other hand, autoregressive models tend to accumulate errors over time and special care must be taken in the training process to mitigate it, thus allowing for practical applications.
The following sections detail the architecture of our discriminator and the learning strategy that enable long sequence generation.

\subsection{Window-based multi-scale discriminator}
\label{subsec:wbmsd}
We use a multi-scale, window-based discriminator network architecture to train the model to generate temporal patterns unfolding over different timescales.
To relieve the burden of training one network per input scale, we follow~\citet{airale2022socialinteractiongan} who achieved this objective using a recurrent network (RNN), considerably simplifying the discriminator architecture.
Here we give a more formal definition of the window-based multi-scale discriminator that is not restricted to RNN variants.
First, let $D_{\textsc{m}}: (x_{t:t+\tau}, \theta) \in \mathbb{R}^{\tau \times d}  \times \mathbb{R}^{d_{\theta}} \mapsto D_{\textsc{m}}(x_{t:t+\tau}; \theta) \in \mathbb{R}$ be a discriminator function parameterized by $\theta$ that operates on sequences of $d$-dimension vectors of arbitrary length $\tau$.
This definition includes RNNs, Transformers~\cite{vaswani2017attention}, and more generally any function enabling pooling in the time axis or processing time steps separately.
We then define the window-based multi-scale discriminator $D_\textsc{s}$ on sequences $x_{0:T}$ of length $T$ as an expectation over evaluations of $D_{\textsc{m}}$ on temporal slices of $x_{0:T}$:
\begin{equation}
    D_{\textsc{s}}(x_{0:T}; \theta) = \mathbb{E}_{\tau, t} [D_{\textsc{m}}(x_{t:t+\tau}; \theta)], \quad t \geq 0, t+\tau \leq T
\end{equation}
where $\tau$ and $t$ are the duration, or equivalently the scale, and starting index of the window.
In practice both $t$ and $\tau$ are sampled from discrete uniform distributions.
The advantage of this framework is that it gives a flexible way to adjust the scales by choosing other distributions on $\tau$.

\newcommand{\xa}{x^{1}}
\newcommand{\xb}{x^{2}}
\subsection{Learning to generate and discriminate joint probability distributions}
\label{subsec:joint}
To mitigate the mode collapse problem, we consider both the generation and discrimination of joint sample distributions.
Let the objective, with generic data points $\xa$ and $\xb$, write (superscript \textsc{j} for joint ground truth / generated distributions): 
\begin{equation}
\mathbb{E}_{\xa, \xb \sim p_{\text{data}}^\textsc{j}}[\log D(\xa, \xb)] + \mathbb{E}_{\xa, \xb \sim p_{\textsc{g}}^{\textsc{j}}}[\log (1 - D(\xa, \xb))]
\end{equation}
This has to be minimized (resp. maximized) w.r.t.\ the parameters of the generator $G$ (resp. the discriminator $D$).
In the case of independent and identically distributed data and enough network capacity, the joint generated distribution converges to the product of the marginal data distributions~\cite{goodfellow2014generative}:
\begin{equation}
\label{eq_d_fact}
p_{\textsc{g}}^{\textsc{j}}(\xa, \xb) = p_{\text{data}}^{\textsc{j}}(\xa, \xb) = p_{\text{data}}(\xa)p_{\text{data}}(\xb)
\end{equation}

If $G$ produces samples independently, then $p_{\textsc{g}}^{\textsc{j}}$ readily factorizes.
This is the setting of~\cite{lin2018pacgan}, which proved useful to reduce mode collapse.
However, if $x$ and $y$ are produced together, then $G$ simply \textit{learns} to factorize.
Both cases lead to the equality of marginal distributions $p_{\textsc{g}} = p_{\text{data}}$, hence the optimization objective of~\citet{goodfellow2014generative} is unaffected.
In the real case scenario of limited network capacity, $p_{\textsc{g}}^{\textsc{j}}$ does not factorize, and hence we argue that if the generation is prone to mode collapse then the overall optimization can benefit from this joint generation process. 
In such cases, it is an easy task for $D$ to identify generated pairs by comparing the two samples, hence driving $G$ to leverage its two inputs to increase the generation diversity.

At test time, a single initial pose is typically provided.
Since the model expects a pair of samples, one strategy consists in providing a transformed version of the reference pose as a second input.
To that end we used random flip, rescaling and translation.
This approach gives a practical way of injecting stochasticity in the generation process (see Section~\ref{subsec:quali}).

\begin{figure*}
\includegraphics[width=0.95\linewidth]{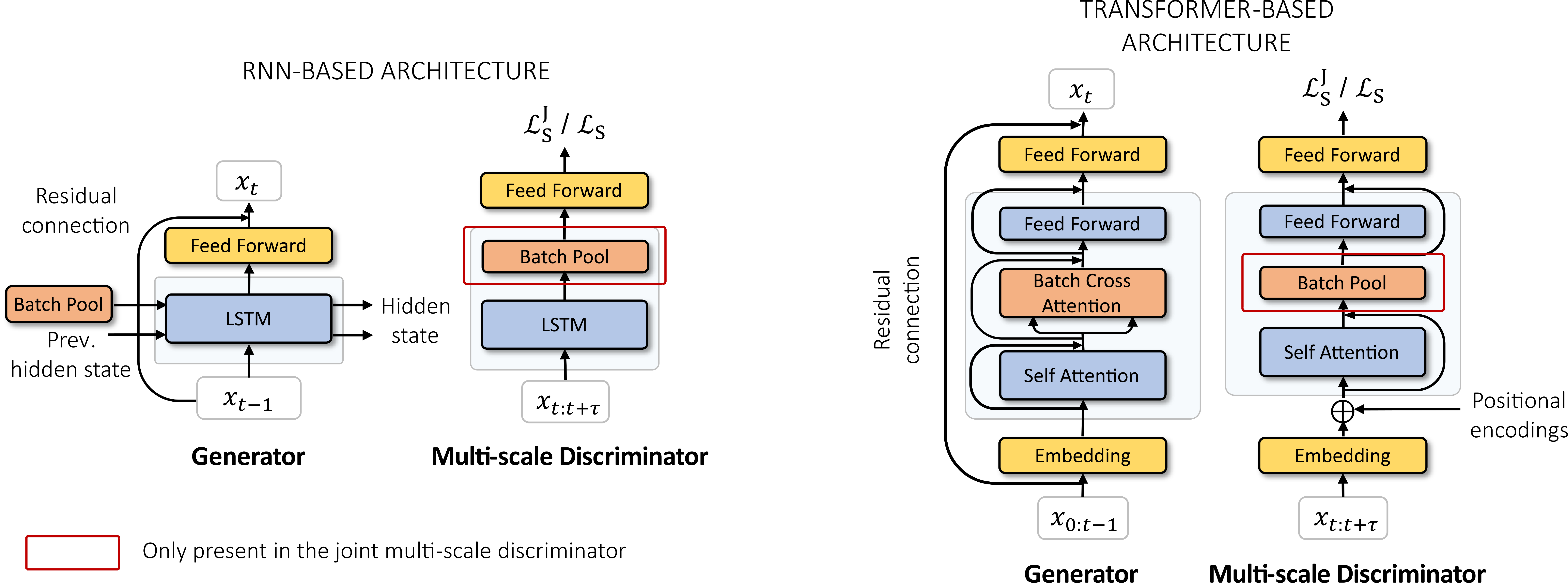}
\centering
\caption{The two architecture variants of the proposed \method{} method.}
\label{fig:architecture}
\end{figure*}

\subsection{Training SUMHo}
\label{subsec:arch_loss}

Following the discussion in \ref{subsec:wbmsd} and \ref{subsec:joint}, we propose to use two window-based multi-scale discriminators on the generated sequences.
The joint discriminator $D_{\textsc{s}}^{\textsc{j}}$ operates on sample pairs, while a second network, $D_{\textsc{s}}$, takes single sequences as input and explicitly enforces the convergence of the marginal distributions $p_{\textsc{g}}$ and $p_{\text{data}}$.
Finally, to complement the sequential losses, we employ a frame discriminator $D_{\textsc{f}}$ to measure the realism of each time step of the produced sequences (see Figure~\ref{fig:overview}).

The loss function is the sum of the three corresponding adversarial losses (joint sequential, sequential, and frame-wise), plus a mean squared error loss $\mathcal{L}_2^{reco}$ that we scale to remain negligible after the first training epochs:

\begin{equation}
\label{eq:loss}
    \mathcal{L} = \underbrace{(\mathcal{L}_{\textsc{s}}^{\textsc{j}} + \mathcal{L}_{\textsc{s}} + \mathcal{L}_{\textsc{f}})}_\text{Adversarial loss} +
    \lambda\mathcal{L}_2^{reco}
\end{equation}

\subsection{Implementation}
\label{subsec:implem}

So far the discussion has not assumed any precise functional form for either the generator or the discriminator network.
Here
we propose two implementations of the \method{} method, based on LSTM and Transformers.
The motivation is to highlight that the provided methodological tools can be relevant beyond a single architecture, as we further discuss in Section~\ref{sec:expe}.
An overview of both proposed variants can be found in Figure~\ref{fig:architecture}.
To account for pairs of inputs, we define a batch-pool operator that acts as a max pooling layer of kernel size 2 along the batch dimension; with the difference that the result is then repeated to preserve the input batch size.

In the LSTM-based generator, the hidden state $h_t$ goes through a batch-pool layer, yielding a pooled vector $p_t$ that is concatenated with the next input to the LSTM.
A multi-layer perceptron is used on $h_t$ to output the landmark positions.
The joint discriminator $D_{\textsc{s}}^{\textsc{j}}$ is composed of a LSTM, a batch-pool layer and a feed forward network; the marginal discriminator $D_{\textsc{s}}$ is similar but without the batch-pool layer. 

In the Transformer generator, pair mixing is done in a multi-head attention layer that takes input pairs stacked in the batch dimension as queries, and the same pairs in reversed order as keys and values.
This way, each sample in a pair can attend to the history of the other sample.
This layer is labelled batch-cross attention.
We do not use positional encoding as we observed no change in performance, while omitting it allows the generation of longer test sequences.
As for the discriminator networks, a batch-pool layer replaces the batch-cross attention in $D_{\textsc{s}}^{\textsc{j}}$ as it only needs to provide a single score per pair.
A learnable class token, prepended to the input sequence, is used to give the final score, as it has been customary for Transformers \cite{devlin-etal-2019-bert}.

\section{Experiments}
\label{sec:expe}

\subsection{Experimental details}

\begin{figure*}
\includegraphics[width=1.0\linewidth]{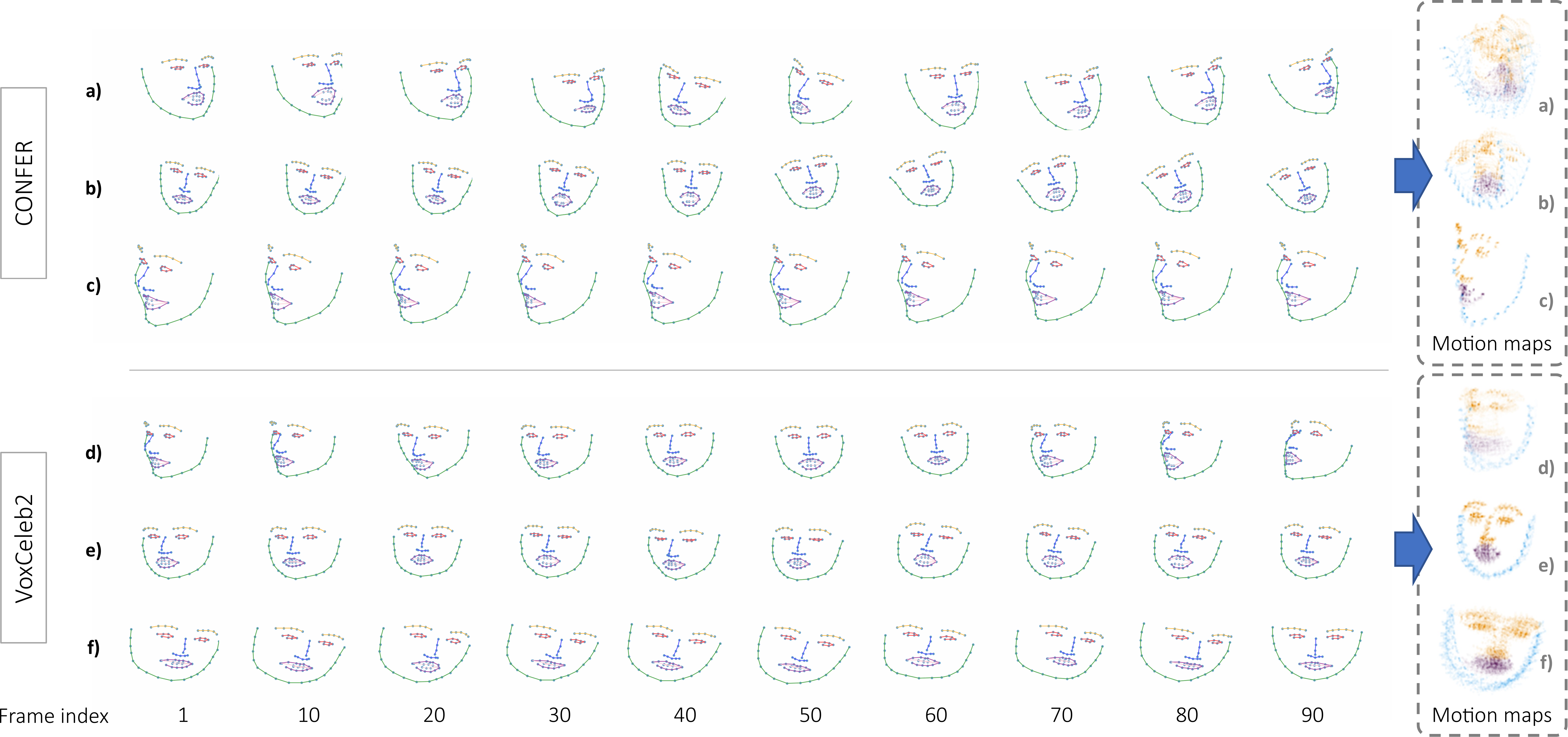}
\centering
\caption{Sample sequences from CONFER and VoxCeleb2 datasets and the associated \textit{motion maps}. Samples featuring little movement produce a very sharp motion map (example c).
The other samples give a good illustration of the differences in dataset preprocessing: head translation is suppressed in VoxCeleb2 sequences that only contain rotations, hence the quasi-static position of noise-tip landmarks in d, e and f. On the contrary, both translation and rotation movements are visible in the motion maps of sample a and b.}
\label{fig:samples_and_motion_maps}
\end{figure*}

We used 1-layer LSTM with hidden size 1024 for all networks in the RNN variant of our method, and a single 1-head self attention block for the Transformer networks. 
In the latter architecture, embedding layers produce 1024 dimensional vectors for the generator and 128 dimensional vectors for the discriminators, i.e. the balance between $G$ and $D$ is mainly controlled by the embedding dimension.
Models were trained on sequences of 40 time steps, and up to 5 observed frames were given as input to the LSTM to stabilize training.
At inference time a single reference frame is provided, and we explore predicting sequences of two different durations, namely 40 and 80 time steps, or respectively $1.6s$ and $3.2s$.

The hinge version of the GAN loss was used in all experiments, and we set $\lambda$ in equation~\ref{eq:loss} to $10^{-2}$.
Networks were trained with Adam optimizers with $\beta_1$ and $\beta_2$ parameters set to 0.5 and 0.999, and with generator and discriminator learning rates set to $2 \times 10^{-5}$ and $1 \times 10^{-5}$ respectively.
Importantly, a step learning rate decay of a factor 10 was applied once performance started to stall, corresponding to roughly 60k iterations for a batch size of 120 ($\sim3000$ epochs for CONFER and $1000$ epochs for our VoxCeleb2 subset).
Training took on average two days on a single Titan RTX GPU.

We investigated concatenating velocities or instantaneous accelerations to landmark positions as input to the generator or the discriminators, expecting that it might help penalizing static sequences produced by $G$.
In practice, we use positions and velocities as inputs to the generator and all three quantities in the discriminator networks.

Experiments were conducted on two audio-visual datasets with upper-body frontal views of different speakers.
\textbf{CONFER}~\cite{georgakis2017conflict} contains 72 video clips of TV debates between two persons, each about 1 minute long.
We pre-processed the data preserving head translations and selected 5 clips as test data featuring persons unseen at training.
Second, we trained on a randomly selected subset from \textbf{VoxCeleb2}~\cite{Chung18b}, leaving 674 video clips corresponding to 10 unseen identities as test set.
In both datasets the video frame rate is 25 fps.

In order to draw robust conclusions despite the inherent variability associated with GAN training, each GAN model was trained three times, such that the results reported in all tables contain both mean values and standard deviations.

\begin{table*}[t]
\caption{Model comparison on the head motion generation task from a single reference frame. FVD and \textit{t}-FID are linked to a fixed sequence length that is reported as subscript. Here all metrics are computed over the 40 last predicted time steps.
}
\label{table_model_compa}
\begin{center}
\resizebox{1.0\textwidth}{!}{
\begin{tabular}{l | ccc | ccc | ccc | ccc}
\toprule
\multicolumn{1}{c|}{Dataset}&\multicolumn{6}{c|}{CONFER~\cite{georgakis2017conflict}} & \multicolumn{6}{c}{VoxCeleb2~\cite{Chung18b}} \\
 \midrule
\multicolumn{1}{c|}{Out length (frames)}&\multicolumn{3}{c|}{40}&\multicolumn{3}{c|}{80}&\multicolumn{3}{c|}{40}&\multicolumn{3}{c}{80}\\
\midrule
Method & $\text{FVD}_{40}$ & $\text{FID}$ & $\text{\textit{t}-FID}_{40}$ &
$\text{FVD}_{40}$ & $\text{FID}$ & $\text{\textit{t}-FID}_{40}$
& $\text{FVD}_{40}$ & $\text{FID}$ & $\text{\textit{t}-FID}_{40}$ &
$\text{FVD}_{40}$ & $\text{FID}$ & $\text{\textit{t}-FID}_{40}$\\
\midrule
HiT-DVAE~\cite{bie2022hit}& $368{\pm 19}$ & $6{\pm 0.4}$ & $130{\pm 7}$ & $764{\pm 35}$ & $50{\pm 2}$ & $157{\pm 12}$ 
& $686{\pm 37}$ & $\textbf{1}{\pm \textbf{0.1}}$ & $167{\pm 4}$ & $644{\pm 27}$ & $\textbf{2}{\pm \textbf{0.1}}$ & $164{\pm 6}$ \\

ACTOR~\cite{petrovich2021action}& $480{\pm 12}$ & $8{\pm 0.3}$ & $147{\pm 3}$ & $667{\pm 20}$ & $9{\pm 0.8}$ & $163{\pm 5}$ 
& $357{\pm 55}$ & $4{\pm 0.5}$ & $78{\pm 9}$ & $431{\pm 26}$ & $5{\pm 2}$ & $145{\pm 21}$\\

$\Delta$-based & $318{\pm 115}$ & $21{\pm 3}$ & $67{\pm 10}$ & $357{\pm 104}$ & $24{\pm 3}$ & $77{\pm 18}$ 
& $386{\pm 32}$ & $48{\pm 6}$ & $89{\pm 4}$ & $518{\pm 48}$ & $60{\pm 30}$ & $112{\pm 31}$\\

$L_2$-only& $480{\pm 42}$ & $10{\pm 3}$ & $133{\pm 2}$ & $777{\pm 54}$ & $21{\pm 3}$ & $159{\pm 6}$ 
& $530{\pm 20}$ & $2{\pm 0.2}$ & $158{\pm 6}$ & $684{\pm 23}$ & $8{\pm 0.8}$ & $172{\pm 9}$\\

\midrule

\method{} - RNN & $\textbf{162}{\pm \textbf{31}}$ & $\textbf{3}{\pm \textbf{0.2}}$ & $\textbf{61}{\pm \textbf{8}}$& $\textbf{147}{\pm \textbf{45}}$ & $8{\pm 2}$ & $\textbf{48}{\pm \textbf{11}}$ 
& $\textbf{76}{\pm \textbf{8}}$ & $3{\pm 0.7}$ & $\textbf{21}{\pm \textbf{3}}$& $\textbf{135}{\pm \textbf{33}}$ & $9{\pm 5}$ & $\textbf{31}{\pm \textbf{7}}$ \\

\method{} - Trans.& $175{\pm 46}$ & $4{\pm 0.7}$ & $67{\pm 12}$ & $169{\pm 33}$ & $\textbf{7}{\pm \textbf{1}}$ & $52{\pm 4}$
& $134{\pm 33}$ & $3{\pm 0.8}$ & $42{\pm 10}$ & $141{\pm 31}$ & $9{\pm 3}$ & $55{\pm 16}$  \\
\bottomrule
\end{tabular}}
\end{center}
\end{table*}

\begin{figure*}
\includegraphics[width=1.0\linewidth]{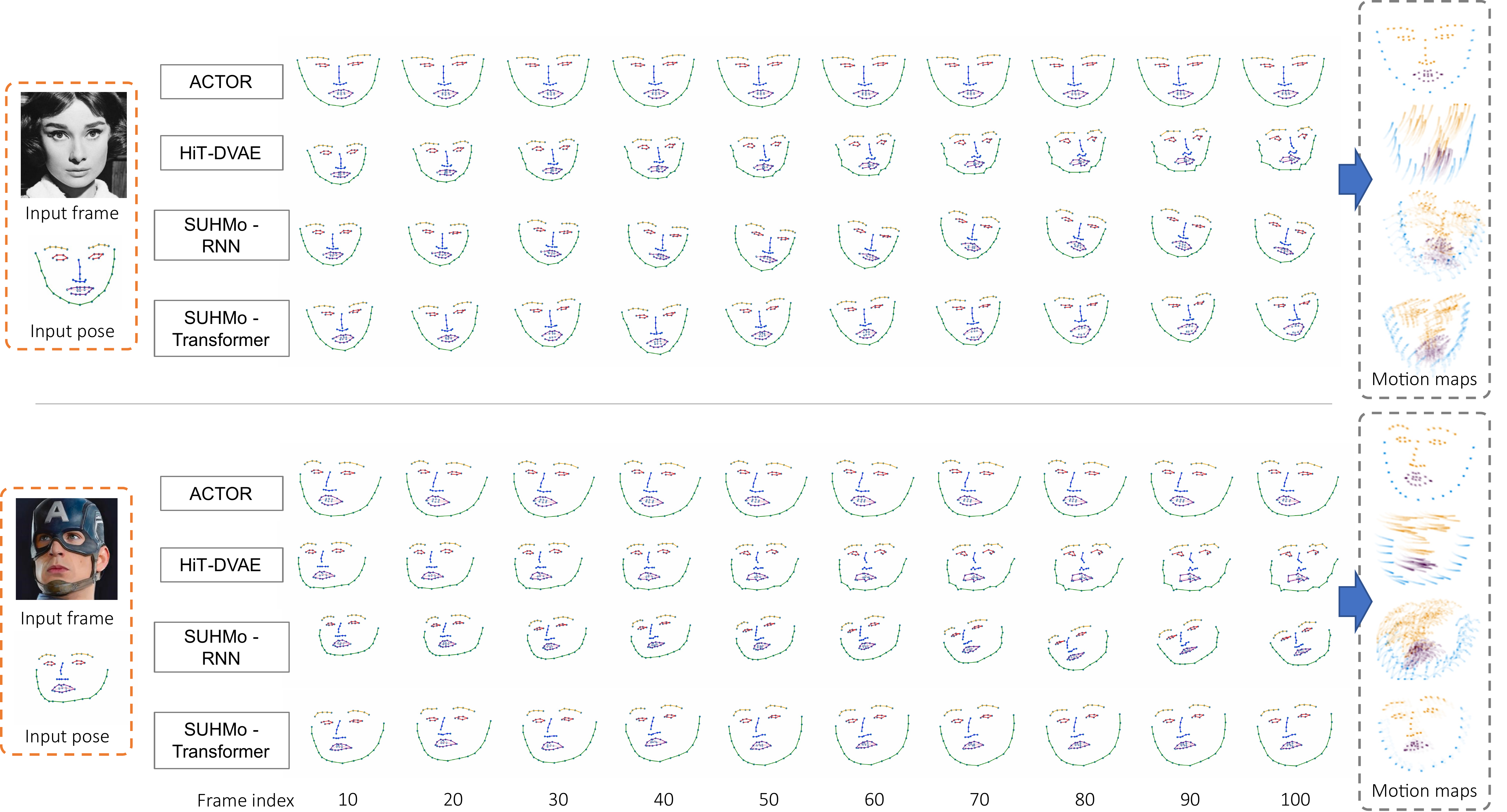}
\centering
\caption{Qualitative evaluation of results from different models on in-the-wild images, and for sequence generation of one hundred frames. Models are trained on CONFER dataset.}
\label{fig:quali_compa}
\vspace{-1.5em}
\end{figure*}

\subsection{Metrics} 

The Fréchet Inception Distance (FID)~\cite{heusel2017gans} and Fréchet Video Distance (FVD)~\cite{unterthiner2018towards} are used to measure the distance of the generated samples to the ground truth data distribution.
While FID gives a score of static face realism, FVD measures the smoothness of the dynamics.
A preliminary rasterization step is applied on landmarks to cast them in the image domain for the inceptionV3~\cite{szegedy2016rethinking} and I3D~\cite{carreira2017quo} networks. 
We also complement the FVD with a second dynamical metric based on a FID measure on \textit{motion maps}, that we use to represent sequences on a single image.
To do so, we compute an exponential moving average centered on the last time frame, thus enforcing a visual correlation between pixel intensity and time step index.
The resulting metric, that is relevant in particular to discriminate sequences with little movement, is coined \textit{t-FID} (t standing for time).
Examples of data samples and their corresponding motion maps are illustrated in Figure~\ref{fig:samples_and_motion_maps}.

\subsection{Models comparison}
\label{subsec:quali}

\paragraph{Quantitative comparison} The performances of \method{} were compared with two state-of-the-art architectures for human pose prediction, \textbf{HiT-DVAE}~\cite{bie2022hit} and \textbf{ACTOR}~\cite{petrovich2021action}.
This task consists in predicted future positions of body joints given a short observed sequence or an action label and is therefore very close to unconditional head motion generation.
One notable difference arises from the training data, which typically contains samples of a predefined set of actions and is therefore explicitly multimodal, contrary to talking head datasets.
To adapt the previous models to our setting, a minimal amount of changes was therefore required.
In particular, we replaced action conditioning in ACTOR by the observed initial frame.
Although we could not directly work with talking head generation methods that require an audio signal for motion prediction, we took inspiration from common practices to build two additional baselines.
The \textbf{$\Delta$-based} model reproduces the \method{}-RNN method, but similarly to~\citet{zhou2020makelttalk} and~\citet{das2020speech} produces displacements from a fixed set of reference points, in this case the initial landmark positions.
\textbf{$L_2$-only} follows a common trend in head motion prediction and relies on a single mean squared error loss.
We evaluate the above models and our two architecture variants on both CONFER and VoxCeleb2, on sequences of duration 40 and 80 frames.
Note that this corresponds to one time and twice the training sequence duration.
Results are reported in table~\ref{table_model_compa}.
\method{} consistently outperforms all other architectures in terms of dynamics quality.
HiT-DVAE and ACTOR attain lower FID values on VoxCeleb2, suggesting slightly sharper faces, but this is at the cost of producing quasi-static sequences, hence the poor FVD and \textit{t}-FID scores (see also next paragraph and Figure~\ref{fig:quali_compa}).
The same is true for models trained with a $L_2$ reconstruction loss, advocating for the use of an adversarial loss to ensure realistic dynamics.
The $\Delta$-based variant produces dynamics of uneven quality, as per the high standard deviations, and the realism of produced faces falls significantly behind, as suggests the higher FID values.
Interestingly, \method{} exhibits very little drift as time stretches and dynamics metrics remain very low, contrary for instance to HiT-DVAE.
We note however that this is an extreme setting for the use of HiT-DVAE in terms of generation over observed length ratio which is typically of the order of 3 to 5 in~\citet{bie2022hit}, whereas here it exceeds 40.

\paragraph{Qualitative evaluation} An illustration of the results of different models on two in-the-wild images is represented in Figure~\ref{fig:quali_compa}, along with the associated motion maps.
It is clear from the observation of motion maps that ACTOR produces very little movement.
HiT-DVAE sequences are likewise almost static, and start drifting after 40 time steps.
\method{} sequences remain sharp aften 100 time steps, suggesting a very limited error accumulation.
These results suggest that despite many similarities in the addressed problems, current human pose prediction models cannot be readily trained on head motion data without suffering a degradation in performance.

\begin{figure*}
\includegraphics[width=1.0\linewidth]{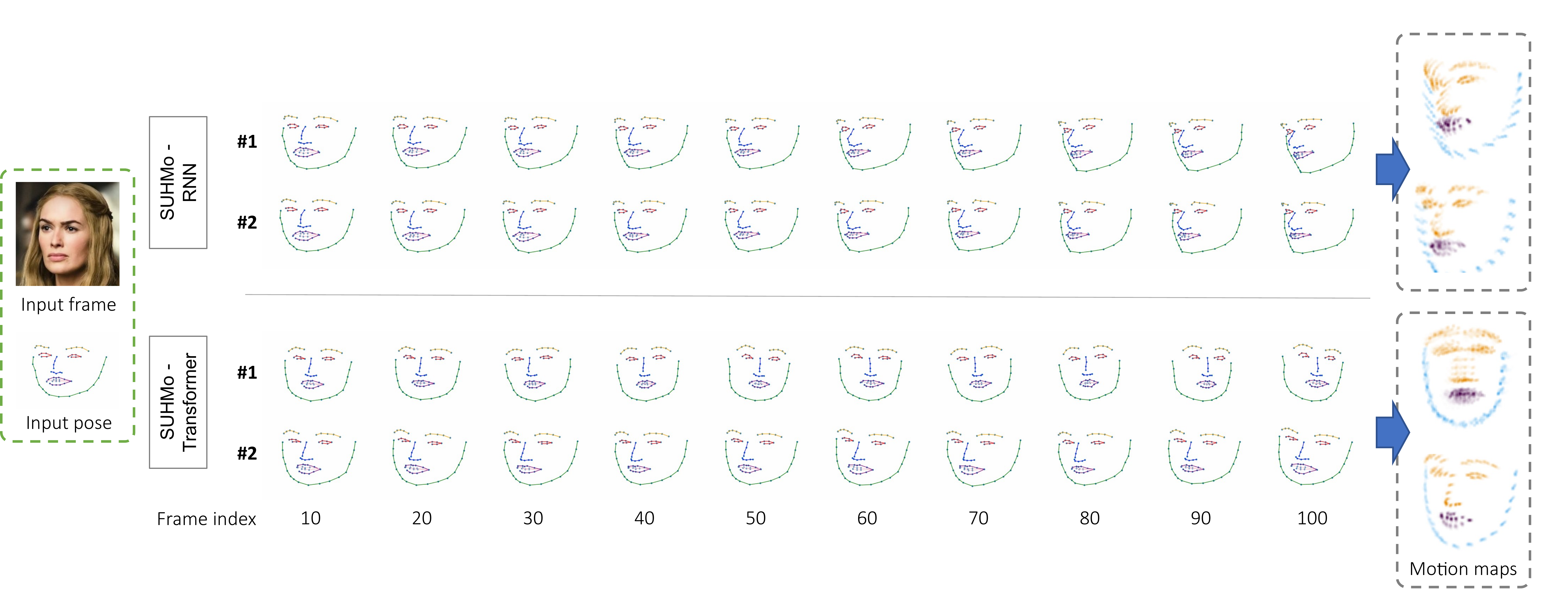}
\centering
\caption{Illustrative examples of diverse results given the same reference pose, for both variants of \method{} method. Models are trained on VoxCeleb2 dataset.}
\label{fig:diversity}
\end{figure*}

An interesting feature of \method{} is that the joint generation allows to produce diverse outputs given the same reference pose.
We illustrate this in Figure~\ref{fig:diversity}.
This is important for many applications that require the ability to generate different outcomes.
These results also show that our training strategy is effective to prevent mode collapse.

\subsection{Ablation study}

\paragraph{Multi-scale discriminator} 
To assess the ability of \method{} to produce realistic patterns over diverse time scales we measure the FVD on motion chunks of 10, 20, and 40 frames, and compare it with a model trained without the window-based multi-scale discriminator (Table~\ref{locald_table_cf}).
Both models were trained to generate sequences of 40 frames and therefore perform on par on this duration.
The benefit of the window-based multi-scale approach however clearly appears on shorter timescales, indicating a finer modeling of high frequency patterns.

\begin{table}[t]
\caption{FVD scores over different sub-sequence lengths, with and without multi-scale window-based discriminator (CONFER). Subscripts indicate the length associated with the metric.}
\label{locald_table_cf}
\begin{center}
\begin{tabular}{l | ccc}
\toprule
Method & $\text{FVD}_{10}$ & $\text{FVD}_{20}$ & $\text{FVD}_{40}$ \\
\midrule
\method{}-RNN & $\textbf{28}{\pm \textbf{8}}$ & $\textbf{35}{\pm \textbf{4}}$ & $162{\pm 31}$ \\
{} w/o multi-scale discriminator & $35{\pm 8}$ & $42{\pm 8}$ & $\textbf{157}{\pm \textbf{21}}$\\

\midrule

\method{}-Transformer& $\textbf{34}{\pm \textbf{9}}$ & $\textbf{40}{\pm \textbf{12}}$ & $\textbf{175}{\pm \textbf{46}}$\\
{} w/o multi-scale discriminator & $57{\pm 6}$ & $60{\pm 12}$ & $236{\pm 58}$\\

\bottomrule
\end{tabular}
\end{center}
\end{table}

\paragraph{Joint generation and discrimination} We tried removing the pair mixing in the generator and the discriminator at turns (Table~\ref{2samples_table_cf}).
Models trained with a standard marginal discriminator ("One-sample D") fall behind in terms of FVD and FID, respectively for the RNN and the Transformer model.
Surprisingly, suppressing the joint generation ("One-sample G") has an even more detrimental effect, visible on FVD and FID for both models.
In addition to its previously known benefits in mode collapse reduction, we observe that working with pairs of samples also helps improving the overall quality of the generated motion sequences in the unconditional generation setting.

\begin{table}[t]
\caption{Two-samples strategy ablation results on CONFER.}
\label{2samples_table_cf}
\begin{center}
\setlength{\tabcolsep}{1mm}
\begin{tabular}{l|ccc|ccc}
\toprule
\method{} variant & \multicolumn{3}{c}{RNN} & \multicolumn{3}{|c}{Transformer} \\
\cmidrule{1-7}
Ablation & $\text{FVD}_{40}$ & $\text{FID}$ & $\text{\textit{t}-FID}_{40}$ 
& $\text{FVD}_{40}$ & $\text{FID}$ & $\text{\textit{t}-FID}_{40}$ \\
\midrule

Full & $\textbf{162}{\pm \textbf{31}}$ & $\textbf{3}{\pm \textbf{0.2}}$ & $\textbf{61}{\pm \textbf{8}}$
& $175{\pm 46}$ & $\textbf{4}{\pm \textbf{0.7}}$ & $67{\pm 12}$\\
One-sample D & $226{\pm 76}$ & $3{\pm 1}$ & $71{\pm 16}$
& $\textbf{162}{\pm \textbf{36}}$ & $7{\pm 1}$ & $\textbf{65}{\pm \textbf{11}}$\\
One-sample G & $222{\pm 23}$ & $5{\pm 0.7}$ & $58{\pm 7}$
& $237{\pm 58}$ & $8{\pm 2}$ & $74{\pm 10}$\\

\bottomrule
\end{tabular}
\end{center}
\end{table}

\section{Conclusion}

In this paper we presented \method{}, an unconditional head motion generation method able to animate a human face over long sequences from a single initial frame in a semantic space.
Our method is based on the autoregressive generation of incremental displacements, or instantaneous velocities, of pairs of samples, and it is trained using a window-based multi-scale discriminator.
We showed that our methodological contributions can accommodate several implementations, consistently outperforming state-of-the-art human pose generation methods and head motion prediction baselines in terms of dynamics quality and pose realism.
In a future work we plan to extend our method and notably assess if it can improve the fidelity of head motion in an audio-conditioned talking head generation setting, which remains an open problem.

\bibliographystyle{abbrvnat}
\bibliography{short_bib}

\begin{thebibliography}{49}
\providecommand{\natexlab}[1]{#1}
\providecommand{\url}[1]{\texttt{#1}}
\expandafter\ifx\csname urlstyle\endcsname\relax
  \providecommand{\doi}[1]{doi: #1}\else
  \providecommand{\doi}{doi: \begingroup \urlstyle{rm}\Url}\fi

\bibitem[Airale et~al.(2022)Airale, Vaufreydaz, and
  Alameda-Pineda]{airale2022socialinteractiongan}
L.~Airale, D.~Vaufreydaz, and X.~Alameda-Pineda.
\newblock Socialinteractiongan: Multi-person interaction sequence generation.
\newblock \emph{IEEE Transactions on Affective Computing}, 2022.

\bibitem[Aliakbarian et~al.(2021)Aliakbarian, Saleh, Petersson, Gould, and
  Salzmann]{aliakbarian2021contextually}
S.~Aliakbarian, F.~Saleh, L.~Petersson, S.~Gould, and M.~Salzmann.
\newblock Contextually plausible and diverse 3d human motion prediction.
\newblock In \emph{Proceedings of the IEEE/CVF International Conference on
  Computer Vision}, pages 11333--11342, 2021.

\bibitem[Arjovsky et~al.(2017)Arjovsky, Chintala, and
  Bottou]{pmlr-v70-arjovsky17a}
M.~Arjovsky, S.~Chintala, and L.~Bottou.
\newblock {W}asserstein generative adversarial networks.
\newblock In D.~Precup and Y.~W. Teh, editors, \emph{Proceedings of the 34th
  International Conference on Machine Learning}, volume~70 of \emph{Proceedings
  of Machine Learning Research}, pages 214--223. PMLR, 06--11 Aug 2017.

\bibitem[Bie et~al.(2022)Bie, Guo, Leglaive, Girin, Moreno-Noguer, and
  Alameda-Pineda]{bie2022hit}
X.~Bie, W.~Guo, S.~Leglaive, L.~Girin, F.~Moreno-Noguer, and X.~Alameda-Pineda.
\newblock Hit-dvae: Human motion generation via hierarchical transformer
  dynamical vae.
\newblock \emph{arXiv preprint arXiv:2204.01565}, 2022.

\bibitem[Carreira and Zisserman(2017)]{carreira2017quo}
J.~Carreira and A.~Zisserman.
\newblock Quo vadis, action recognition? a new model and the kinetics dataset.
\newblock In \emph{proceedings of the IEEE Conference on Computer Vision and
  Pattern Recognition}, pages 6299--6308, 2017.

\bibitem[Che et~al.(2016)Che, Li, Jacob, Bengio, and Li]{che2016mode}
T.~Che, Y.~Li, A.~P. Jacob, Y.~Bengio, and W.~Li.
\newblock Mode regularized generative adversarial networks.
\newblock \emph{arXiv preprint arXiv:1612.02136}, 2016.

\bibitem[Chen et~al.(2019)Chen, Maddox, Duan, and Xu]{chen2019hierarchical}
L.~Chen, R.~K. Maddox, Z.~Duan, and C.~Xu.
\newblock Hierarchical cross-modal talking face generation with dynamic
  pixel-wise loss.
\newblock In \emph{Proceedings of the IEEE/CVF conference on computer vision
  and pattern recognition}, pages 7832--7841, 2019.

\bibitem[Chen et~al.(2020)Chen, Cui, Liu, Li, Kou, Xu, and Xu]{chen2020talking}
L.~Chen, G.~Cui, C.~Liu, Z.~Li, Z.~Kou, Y.~Xu, and C.~Xu.
\newblock Talking-head generation with rhythmic head motion.
\newblock In \emph{European Conference on Computer Vision}, pages 35--51.
  Springer, 2020.

\bibitem[Chen et~al.(2016)Chen, Duan, Houthooft, Schulman, Sutskever, and
  Abbeel]{chen2016infogan}
X.~Chen, Y.~Duan, R.~Houthooft, J.~Schulman, I.~Sutskever, and P.~Abbeel.
\newblock Infogan: Interpretable representation learning by information
  maximizing generative adversarial nets.
\newblock \emph{Advances in neural information processing systems}, 29, 2016.

\bibitem[Chung et~al.(2018)Chung, Nagrani, and Zisserman]{Chung18b}
J.~S. Chung, A.~Nagrani, and A.~Zisserman.
\newblock Voxceleb2: Deep speaker recognition.
\newblock In \emph{INTERSPEECH}, 2018.

\bibitem[Das et~al.(2020)Das, Biswas, Sinha, and Bhowmick]{das2020speech}
D.~Das, S.~Biswas, S.~Sinha, and B.~Bhowmick.
\newblock Speech-driven facial animation using cascaded gans for learning of
  motion and texture.
\newblock In \emph{European conference on computer vision}, pages 408--424.
  Springer, 2020.

\bibitem[Devlin et~al.(2019)Devlin, Chang, Lee, and
  Toutanova]{devlin-etal-2019-bert}
J.~Devlin, M.-W. Chang, K.~Lee, and K.~Toutanova.
\newblock {BERT}: Pre-training of deep bidirectional transformers for language
  understanding.
\newblock In \emph{NAACL}, 2019.

\bibitem[Fan et~al.(2022)Fan, Lin, Saito, Wang, and Komura]{fan2022faceformer}
Y.~Fan, Z.~Lin, J.~Saito, W.~Wang, and T.~Komura.
\newblock Faceformer: Speech-driven 3d facial animation with transformers.
\newblock In \emph{Proceedings of the IEEE/CVF Conference on Computer Vision
  and Pattern Recognition}, pages 18770--18780, 2022.

\bibitem[Georgakis et~al.(2017)Georgakis, Panagakis, Zafeiriou, and
  Pantic]{georgakis2017conflict}
C.~Georgakis, Y.~Panagakis, S.~Zafeiriou, and M.~Pantic.
\newblock The conflict escalation resolution (confer) database.
\newblock \emph{Image and Vision Computing}, 65:\penalty0 37--48, 2017.

\bibitem[Goodfellow et~al.(2014)Goodfellow, Pouget-Abadie, Mirza, Xu,
  Warde-Farley, Ozair, Courville, and Bengio]{goodfellow2014generative}
I.~Goodfellow, J.~Pouget-Abadie, M.~Mirza, B.~Xu, D.~Warde-Farley, S.~Ozair,
  A.~Courville, and Y.~Bengio.
\newblock Generative adversarial nets.
\newblock In \emph{Advances in Neural Information Processing Systems
  (NeurIPS)}, pages 2672--2680, 2014.

\bibitem[Greenwood et~al.(2018)Greenwood, Matthews, and
  Laycock]{greenwood2018joint}
D.~Greenwood, I.~Matthews, and S.~Laycock.
\newblock Joint learning of facial expression and head pose from speech.
\newblock Interspeech, 2018.

\bibitem[Gupta et~al.(2018)Gupta, Johnson, Fei-Fei, Savarese, and
  Alahi]{gupta2018social}
A.~Gupta, J.~Johnson, L.~Fei-Fei, S.~Savarese, and A.~Alahi.
\newblock Social gan: Socially acceptable trajectories with generative
  adversarial networks.
\newblock In \emph{Proceedings of the IEEE Conference on Computer Vision and
  Pattern Recognition (CVPR)}, pages 2255--2264, 2018.

\bibitem[Heusel et~al.(2017)Heusel, Ramsauer, Unterthiner, Nessler, and
  Hochreiter]{heusel2017gans}
M.~Heusel, H.~Ramsauer, T.~Unterthiner, B.~Nessler, and S.~Hochreiter.
\newblock Gans trained by a two time-scale update rule converge to a local nash
  equilibrium.
\newblock \emph{Advances in Neural Information Processing Systems (NeurIPS)},
  30:\penalty0 6626--6637, 2017.

\bibitem[Isola et~al.(2017)Isola, Zhu, Zhou, and Efros]{isola2017image}
P.~Isola, J.-Y. Zhu, T.~Zhou, and A.~A. Efros.
\newblock Image-to-image translation with conditional adversarial networks.
\newblock In \emph{Proceedings of the IEEE conference on computer vision and
  pattern recognition}, pages 1125--1134, 2017.

\bibitem[Ji et~al.(2022)Ji, Zhou, Wang, Wu, Wu, Xu, and Cao]{ji2022eamm}
X.~Ji, H.~Zhou, K.~Wang, Q.~Wu, W.~Wu, F.~Xu, and X.~Cao.
\newblock Eamm: One-shot emotional talking face via audio-based emotion-aware
  motion model.
\newblock \emph{arXiv preprint arXiv:2205.15278}, 2022.

\bibitem[Karras et~al.(2017)Karras, Aila, Laine, Herva, and
  Lehtinen]{karras2017audio}
T.~Karras, T.~Aila, S.~Laine, A.~Herva, and J.~Lehtinen.
\newblock Audio-driven facial animation by joint end-to-end learning of pose
  and emotion.
\newblock \emph{ACM Transactions on Graphics (TOG)}, 36\penalty0 (4):\penalty0
  1--12, 2017.

\bibitem[Kumar et~al.(2019)Kumar, Kumar, de~Boissiere, Gestin, Teoh, Sotelo,
  de~Br{\'e}bisson, Bengio, and Courville]{kumar2019melgan}
K.~Kumar, R.~Kumar, T.~de~Boissiere, L.~Gestin, W.~Z. Teoh, J.~Sotelo,
  A.~de~Br{\'e}bisson, Y.~Bengio, and A.~C. Courville.
\newblock Melgan: Generative adversarial networks for conditional waveform
  synthesis.
\newblock \emph{Advances in neural information processing systems}, 32, 2019.

\bibitem[Kundu et~al.(2019)Kundu, Gor, and Babu]{kundu2019bihmp}
J.~N. Kundu, M.~Gor, and R.~V. Babu.
\newblock Bihmp-gan: Bidirectional 3d human motion prediction gan.
\newblock In \emph{Proceedings of the AAAI conference on artificial
  intelligence}, volume~33, pages 8553--8560, 2019.

\bibitem[Lin and Amer(2018)]{lin2018human}
X.~Lin and M.~R. Amer.
\newblock Human motion modeling using dvgans.
\newblock \emph{arXiv preprint arXiv:1804.10652}, 2018.

\bibitem[Lin et~al.(2018)Lin, Khetan, Fanti, and Oh]{lin2018pacgan}
Z.~Lin, A.~Khetan, G.~Fanti, and S.~Oh.
\newblock Pacgan: The power of two samples in generative adversarial networks.
\newblock \emph{Advances in neural information processing systems}, 31, 2018.

\bibitem[Martinez et~al.(2017)Martinez, Black, and Romero]{martinez2017human}
J.~Martinez, M.~J. Black, and J.~Romero.
\newblock On human motion prediction using recurrent neural networks.
\newblock In \emph{Proceedings of the IEEE conference on computer vision and
  pattern recognition}, pages 2891--2900, 2017.

\bibitem[Meshry et~al.(2021)Meshry, Suri, Davis, and
  Shrivastava]{meshry2021learned}
M.~Meshry, S.~Suri, L.~S. Davis, and A.~Shrivastava.
\newblock Learned spatial representations for few-shot talking-head synthesis.
\newblock In \emph{Proceedings of the IEEE/CVF International Conference on
  Computer Vision}, pages 13829--13838, 2021.

\bibitem[Morrison et~al.(2022)Morrison, Kumar, Kumar, Seetharaman, Courville,
  and Bengio]{morrison2022chunked}
M.~Morrison, R.~Kumar, K.~Kumar, P.~Seetharaman, A.~Courville, and Y.~Bengio.
\newblock Chunked autoregressive {GAN} for conditional waveform synthesis.
\newblock In \emph{International Conference on Learning Representations}, 2022.
\newblock URL \url{https://openreview.net/forum?id=v3aeIsY_vVX}.

\bibitem[Petrovich et~al.(2021)Petrovich, Black, and
  Varol]{petrovich2021action}
M.~Petrovich, M.~J. Black, and G.~Varol.
\newblock Action-conditioned 3d human motion synthesis with transformer vae.
\newblock In \emph{Proceedings of the IEEE/CVF International Conference on
  Computer Vision}, pages 10985--10995, 2021.

\bibitem[Stypu{\l}kowski et~al.(2023)Stypu{\l}kowski, Vougioukas, He,
  Zi{\k{e}}ba, Petridis, and Pantic]{stypulkowski2023diffused}
M.~Stypu{\l}kowski, K.~Vougioukas, S.~He, M.~Zi{\k{e}}ba, S.~Petridis, and
  M.~Pantic.
\newblock Diffused heads: Diffusion models beat gans on talking-face
  generation.
\newblock \emph{arXiv preprint arXiv:2301.03396}, 2023.

\bibitem[Suwajanakorn et~al.(2017)Suwajanakorn, Seitz, and
  Kemelmacher-Shlizerman]{suwajanakorn2017synthesizing}
S.~Suwajanakorn, S.~M. Seitz, and I.~Kemelmacher-Shlizerman.
\newblock Synthesizing obama: learning lip sync from audio.
\newblock \emph{ACM Transactions on Graphics (ToG)}, 36\penalty0 (4):\penalty0
  1--13, 2017.

\bibitem[Szegedy et~al.(2016)Szegedy, Vanhoucke, Ioffe, Shlens, and
  Wojna]{szegedy2016rethinking}
C.~Szegedy, V.~Vanhoucke, S.~Ioffe, J.~Shlens, and Z.~Wojna.
\newblock Rethinking the inception architecture for computer vision.
\newblock In \emph{Proceedings of the IEEE conference on computer vision and
  pattern recognition}, pages 2818--2826, 2016.

\bibitem[Taylor et~al.(2017)Taylor, Kim, Yue, Mahler, Krahe, Rodriguez,
  Hodgins, and Matthews]{taylor2017deep}
S.~Taylor, T.~Kim, Y.~Yue, M.~Mahler, J.~Krahe, A.~G. Rodriguez, J.~Hodgins,
  and I.~Matthews.
\newblock A deep learning approach for generalized speech animation.
\newblock \emph{ACM Transactions on Graphics (TOG)}, 36\penalty0 (4):\penalty0
  1--11, 2017.

\bibitem[Unterthiner et~al.(2018)Unterthiner, van Steenkiste, Kurach, Marinier,
  Michalski, and Gelly]{unterthiner2018towards}
T.~Unterthiner, S.~van Steenkiste, K.~Kurach, R.~Marinier, M.~Michalski, and
  S.~Gelly.
\newblock Towards accurate generative models of video: A new metric \&
  challenges.
\newblock \emph{arXiv preprint arXiv:1812.01717}, 2018.

\bibitem[Vaswani et~al.(2017)Vaswani, Shazeer, Parmar, Uszkoreit, Jones, Gomez,
  Kaiser, and Polosukhin]{vaswani2017attention}
A.~Vaswani, N.~Shazeer, N.~Parmar, J.~Uszkoreit, L.~Jones, A.~N. Gomez,
  {\L}.~Kaiser, and I.~Polosukhin.
\newblock Attention is all you need.
\newblock \emph{Advances in neural information processing systems}, 30, 2017.

\bibitem[Villegas et~al.(2017)Villegas, Yang, Zou, Sohn, Lin, and
  Lee]{villegas2017learning}
R.~Villegas, J.~Yang, Y.~Zou, S.~Sohn, X.~Lin, and H.~Lee.
\newblock Learning to generate long-term future via hierarchical prediction.
\newblock In \emph{international conference on machine learning}, pages
  3560--3569. PMLR, 2017.

\bibitem[Vondrick et~al.(2016)Vondrick, Pirsiavash, and
  Torralba]{vondrick2016generating}
C.~Vondrick, H.~Pirsiavash, and A.~Torralba.
\newblock Generating videos with scene dynamics.
\newblock In \emph{Advances in Neural Information Processing Systems
  (NeurIPS)}, pages 613--621, 2016.

\bibitem[Vougioukas et~al.(2020)Vougioukas, Petridis, and
  Pantic]{vougioukas2020realistic}
K.~Vougioukas, S.~Petridis, and M.~Pantic.
\newblock Realistic speech-driven facial animation with gans.
\newblock \emph{International Journal of Computer Vision}, 128\penalty0
  (5):\penalty0 1398--1413, 2020.

\bibitem[Wang et~al.(2021)Wang, Li, Ding, Fan, and Yu]{Wang2021Audio2HeadAO}
S.~Wang, L.~Li, Y.~Ding, C.~Fan, and X.~Yu.
\newblock Audio2head: Audio-driven one-shot talking-head generation with
  natural head motion.
\newblock In \emph{IJCAI}, 2021.

\bibitem[Wang et~al.(2022)Wang, Li, Ding, and Yu]{wang2022one}
S.~Wang, L.~Li, Y.~Ding, and X.~Yu.
\newblock One-shot talking face generation from single-speaker audio-visual
  correlation learning.
\newblock In \emph{Proceedings of the AAAI Conference on Artificial
  Intelligence}, volume~36, pages 2531--2539, 2022.

\bibitem[Wang et~al.(2018)Wang, Liu, Zhu, Tao, Kautz, and
  Catanzaro]{wang2018high}
T.-C. Wang, M.-Y. Liu, J.-Y. Zhu, A.~Tao, J.~Kautz, and B.~Catanzaro.
\newblock High-resolution image synthesis and semantic manipulation with
  conditional gans.
\newblock In \emph{Proceedings of the IEEE conference on computer vision and
  pattern recognition}, pages 8798--8807, 2018.

\bibitem[Yi et~al.(2020)Yi, Ye, Zhang, Bao, and Liu]{yi2020audio}
R.~Yi, Z.~Ye, J.~Zhang, H.~Bao, and Y.-J. Liu.
\newblock Audio-driven talking face video generation with learning-based
  personalized head pose.
\newblock \emph{arXiv preprint arXiv:2002.10137}, 2020.

\bibitem[Yu et~al.(2020)Yu, Yu, Li, and Ling]{yu2020multimodal}
L.~Yu, J.~Yu, M.~Li, and Q.~Ling.
\newblock Multimodal inputs driven talking face generation with
  spatial--temporal dependency.
\newblock \emph{IEEE Transactions on Circuits and Systems for Video
  Technology}, 31\penalty0 (1):\penalty0 203--216, 2020.

\bibitem[Zakharov et~al.(2019)Zakharov, Shysheya, Burkov, and
  Lempitsky]{zakharov2019few}
E.~Zakharov, A.~Shysheya, E.~Burkov, and V.~Lempitsky.
\newblock Few-shot adversarial learning of realistic neural talking head
  models.
\newblock In \emph{Proceedings of the IEEE International Conference on Computer
  Vision (ICCV)}, pages 9459--9468, 2019.

\bibitem[Zakharov et~al.(2020)Zakharov, Ivakhnenko, Shysheya, and
  Lempitsky]{zakharov2020fast}
E.~Zakharov, A.~Ivakhnenko, A.~Shysheya, and V.~Lempitsky.
\newblock Fast bi-layer neural synthesis of one-shot realistic head avatars.
\newblock In \emph{European Conference on Computer Vision}, pages 524--540.
  Springer, 2020.

\bibitem[Zhao et~al.(2021)Zhao, Wu, and Guo]{zhao2021sparse}
R.~Zhao, T.~Wu, and G.~Guo.
\newblock Sparse to dense motion transfer for face image animation.
\newblock In \emph{Proceedings of the IEEE/CVF International Conference on
  Computer Vision}, pages 1991--2000, 2021.

\bibitem[Zhou et~al.(2019)Zhou, Liu, Liu, Luo, and Wang]{zhou2019talking}
H.~Zhou, Y.~Liu, Z.~Liu, P.~Luo, and X.~Wang.
\newblock Talking face generation by adversarially disentangled audio-visual
  representation.
\newblock In \emph{Proceedings of the AAAI conference on artificial
  intelligence}, volume~33, pages 9299--9306, 2019.

\bibitem[Zhou et~al.(2021)Zhou, Sun, Wu, Loy, Wang, and Liu]{zhou2021pose}
H.~Zhou, Y.~Sun, W.~Wu, C.~C. Loy, X.~Wang, and Z.~Liu.
\newblock Pose-controllable talking face generation by implicitly modularized
  audio-visual representation.
\newblock In \emph{Proceedings of the IEEE/CVF conference on computer vision
  and pattern recognition}, pages 4176--4186, 2021.

\bibitem[Zhou et~al.(2020)Zhou, Han, Shechtman, Echevarria, Kalogerakis, and
  Li]{zhou2020makelttalk}
Y.~Zhou, X.~Han, E.~Shechtman, J.~Echevarria, E.~Kalogerakis, and D.~Li.
\newblock Makelttalk: speaker-aware talking-head animation.
\newblock \emph{ACM Transactions on Graphics (TOG)}, 39\penalty0 (6):\penalty0
  1--15, 2020.

\end{thebibliography}

\end{document}